\documentclass{article}

\usepackage[final]{corl_2019} 

\usepackage{ulem}
\usepackage{graphicx}
\usepackage{subcaption}
\usepackage{algorithm}
\usepackage{algorithmic}
\usepackage{amsmath}
\usepackage{amssymb}
\usepackage{enumitem}

\newcommand{\Skip}[1]{}

\newcommand{\eg}{e.g.\ }
\newcommand{\ie}{i.e.\ }

\newcommand{\figref}[1]{Figure~\ref{fig:#1}}
\newcommand{\algref}[1]{Algorithm~\ref{alg:#1}}
\renewcommand{\eqref}[1]{Equation~(\ref{eq:#1})}

\newcommand{\tabref}[1]{Table~\ref{tab:#1}}

\title{To Follow or not to Follow: \\ Selective Imitation Learning from Observations}

\author{
  Youngwoon Lee, Edward S. Hu, Zhengyu Yang, and Joseph J. Lim\\
  Department of Computer Science\\
  University of Southern California\\
  \texttt{\{ lee504, hues, yang765, limjj \}@usc.edu} \\
}

\begin{document}
\maketitle


\begin{abstract}
Learning from demonstrations is a useful way to transfer a skill from one agent to another. While most imitation learning methods aim to mimic an expert skill by following the demonstration step-by-step, imitating every step in the demonstration often becomes infeasible when the learner and its environment are different from the demonstration. In this paper, we propose a method that can imitate a demonstration composed solely of observations, which may not be reproducible with the current agent. Our method, dubbed selective imitation learning from observations (SILO), selects reachable states in the demonstration and learns how to reach the selected states. Our experiments on both simulated and real robot environments show that our method reliably performs a new task by following a demonstration. Videos and code are available at \url{https://clvrai.com/silo}.
\end{abstract}

\keywords{imitation learning, hierarchical reinforcement learning, deep learning} 


\section{Introduction}

Humans often learn skills such as tying a necktie or cooking new dishes by simply watching a video and following it step-by-step. When humans learn a skill from a demonstration, there are two challenges that must be overcome. First, we do not have direct access to the demonstrator's actions, so we must observe the behavior of the demonstrator and deduce the actions necessary to realize the demonstration. The second challenge arises when the imitator cannot strictly imitate the demonstration because of differing physical capabilities or environmental constraints (\eg new obstacles) between the imitator and the demonstrator. For example, we often need to decide which steps of the demonstration to follow through or to ignore and skip with respect to our own capabilities.

Can machines similarly learn new skills by following a demonstration taking into account their own physical capabilities and environmental discrepancies? 
Learning from demonstrations~\citep{schaal1997learning} (LfD) has been actively studied as an effective and convenient approach to teach robots complex skills. 
However, typical LfD methods assume access to the underlying actions.
While \citet{sermanet2018time, liu2018imitation, peng2018sfv} aim to imitate human demonstrations without access to the underlying actions, they assume that a demonstration can be temporally aligned with the agent's actions.
This assumption does not hold when the demonstrator and learner have different physical abilities and environments (\eg learning from human demonstrations).
In such a case, instead of strictly following a demonstration step-by-step, an agent needs to adaptively select the frames that are feasible to follow (i.e. account for environmental constraints and temporal misalignment) and then imitate.

In this paper, we propose a method that can imitate a demonstration which may be infeasible to precisely follow due to the differences in the agent capabilities or the environment setups.
Our method, selective imitation learning from observations (SILO), learns to follow the demonstration by aligning the timescale of the demonstrator and learner and skipping infeasible parts of the demonstration. 
This problem requires two mechanisms, one to discriminate between feasible and infeasible frames in the demonstration and another to deduce the actions required to reach the feasible frames.
To achieve this, SILO has a hierarchy of policies: a meta policy and a low-level policy.
Specifically, the meta policy selects a frame from the demonstration as a sub-goal for the low-level policy. This effectively controls the pace of imitation for the current agent and decides which parts of the demonstration to ignore. On the other hand, the low-level policy takes the sub-goal as an input from the meta policy, and generates a sequence of actions until the agent reaches the chosen sub-goal. This process does not require action labels, and thus the low-level policy learns how to achieve the frame from the demonstration instead of merely imitating the expert action. 
Once the agent achieves the sub-goal, the meta policy picks the next sub-goal. 
An agent can imitate the whole demonstration by repeating this process. 
By maximizing the number of achieved sub-goals, the meta policy learns to imitate the behavior as closely as possible while having the flexibility to deviate from frames that are impossible to follow.

The contributions of this paper include a concept of learning to selectively imitate a demonstration and a hierarchical approach to learn from a temporally unaligned demonstration of observations.
The proposed method is evaluated on \textsc{Obstacle push} and \textsc{Pick-and-place} with a Sawyer robot in simulation, \textsc{Furniture assembly} environment~\citep{lee2019ikea} in simulation, and \textsc{Obstacle push (Real)} with a Sawyer robot in the real physical world.
We show that SILO works with both raw state and estimated abstract state (\eg predicted 3d object position).
The extensive experiments show that SILO effectively learns to perform a complex task described in a single demonstration.

\begin{figure}[!t]
    \centering
    \includegraphics[width=.95\linewidth]{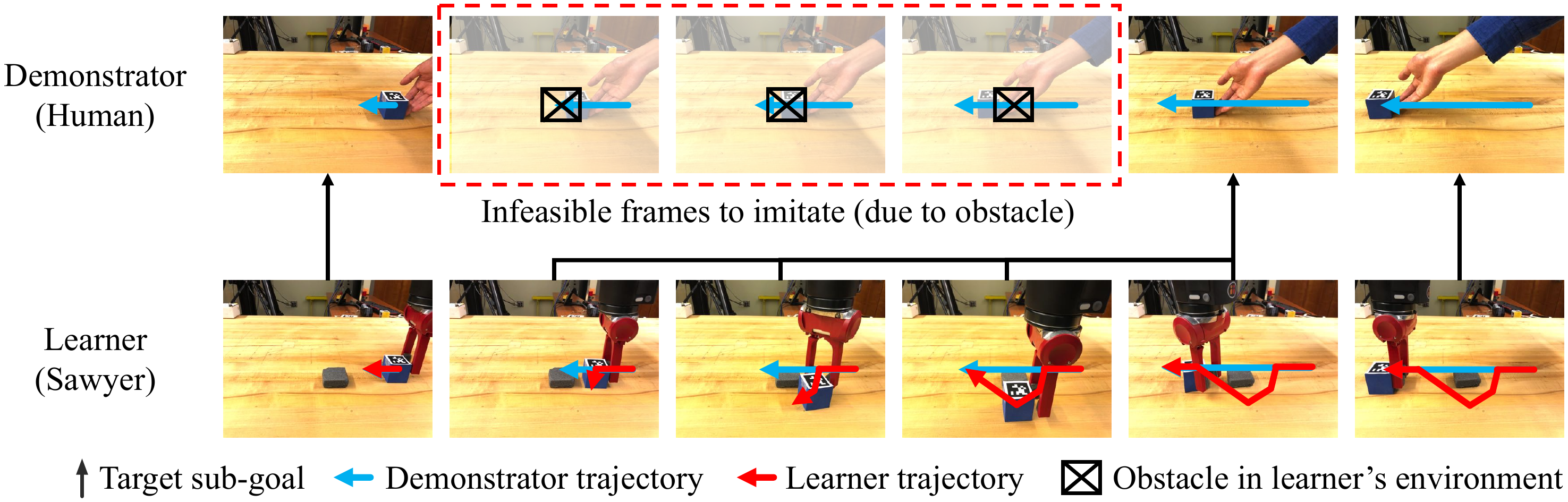}
    \caption{
    Our method learns a task by following a demonstration flexibly. The top row shows a human demonstration from which a learner (Sawyer) tries to imitate. 
    A demonstration is often misaligned with a learner's action since each individual has different physical constraints and environmental limitations. 
    The demonstration on the top row contains some unreachable frames (red box) with respect to the learner due to an obstacle. Hence, the learner ignores those frames and imitates the following frame directly. 
    }
    \label{fig:teaser}
\end{figure}

\section{Related Work}

Imitation learning aims to teach machines a desired behavior by imitating expert demonstrations.
Typical imitation learning methods assume access to state-action pairs, but acquiring expert actions is expensive.
To imitate new demonstrations without action labels, meta-learning approaches~\citep{duan2017one-shot, finn2017one, yu2018aone, yu2018bone} adapt a policy to the given tasks and program synthesis methods~\citep{sun2018neural} generate a new program for the given demonstrations.
However, these approaches require many pairs of expert and robot demonstrations during the training stage as well as action supervision.

Instead of learning from expensive demonstrations of state-action pairs, learning from observations, especially human demonstration videos, has been recently proposed.
Perceptual reward~\citep{sermanet2017rewards} can be learned from a set of human demonstrations and used for reinforcement learning. 
This method cannot handle multi-modal tasks, such as moving a box to the left and to the right since the learned reward predicts whether the current observation is close to or deviates from a common behavior in demonstrations. 
However, our method can capture multi-modal behaviors by conditioning on the given demonstration.
Estimated pose~\citep{peng2018sfv}, visual embeddings learned from time-contrastive networks~\citep{sermanet2018time} and context translation~\citep{liu2018imitation} are used to provide dense reward signals by comparing the current observation and the demonstration frame. 
These approaches require a demonstration to be temporally aligned with an agent's execution, whereas our proposed method learns to align the current observation to the demonstration so that it can go slower or faster than the demonstration, and even skip some infeasible states.

Some tasks can be effectively described by a single goal observation, such as object manipulation~\cite{deguchi1999image,watter2015embed,nair2017combining, lee2019composing}, visuomotor control~\cite{finn2016deep}, and visual navigation~\cite{zhu2017target,pathak*2018zeroshot}.
However, composite and temporally extended tasks, such as dancing and obstacle course, are described by a sequence of observations rather than a single goal. 
Our proposed method aims to both reach the goal state and mimic the expert behavior as closely as possible.

\section{Approach}

\begin{figure*}[t]
\centering
    \includegraphics[width=.95\textwidth]{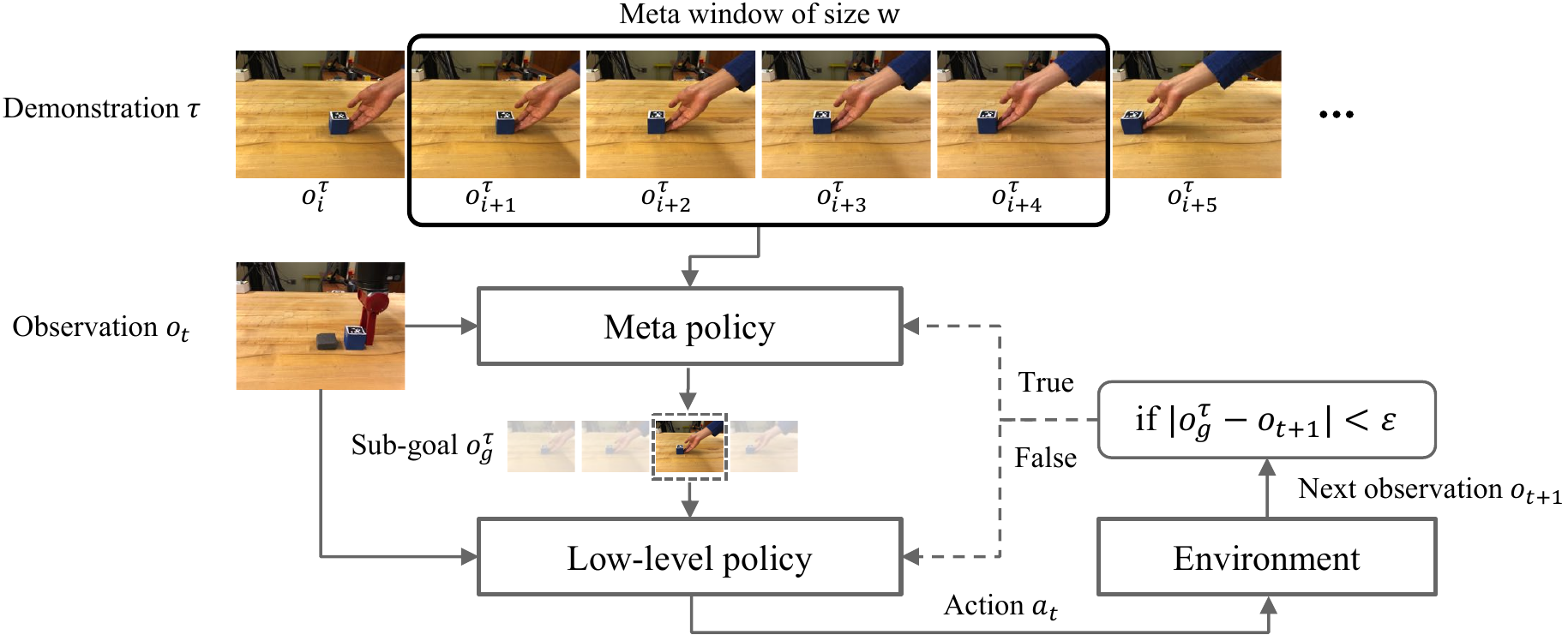}
    \caption{
        Our hierarchical RL framework for selective imitation learning from observations. 
        To follow the demonstration $\tau$, a meta policy first picks a state $o^\tau_g \in \tau$ as a sub-goal for a low-level policy that is reachable from the current observation $o_t$.
        Then, the low-level policy deduces an action $a_t$ based on the observation $o_t$ and the sub-goal $o^\tau_g$ to reach the sub-goal. 
        After the execution of $a_t$, if the agent reaches the sub-goal (\ie $|o^\tau_g - o_{t+1}| < \epsilon$), the meta policy will pick the next sub-goal. Otherwise, the low-level policy repeatedly generates actions until it reaches the sub-goal.  
        \label{fig:model}
    }
\end{figure*}

In this paper, we address the problem of imitating a given demonstration that only consists of observations. 
Naively following a demonstration does not work when the demonstration contains a state that is unreachable by the learner agent or if the demonstrator has different physical dynamics from the learner, such as movement speed or degrees of freedom.
We propose selective imitation learning from observations (SILO), a hierarchical RL framework that learns to find a feasible demonstration state from the current state and move the agent to the chosen state.

\subsection{Problem Formulation}
\label{sec:problem}

Our goal is to imitate a demonstration $\tau=\{o^\tau_1, o^\tau_2, \dots, o^\tau_T\}$ which is a sequence of observations, where the underlying expert actions are not accessible.
In this setup, an agent should learn its policy by interacting with an environment since it cannot learn to predict actions that mimic expert actions as in behavioral cloning methods~\citep{ross2011dagger} due to the lack of action supervision.
Moreover, in our setup, no explicit reward function is provided, which means the goal of an agent is to mimic a demonstration as closely as possible. 
Instead of having an explicit reward function, we directly compare the current state and the demonstration state to indicate how closely the agent is following the demonstration. 
Since many different embeddings can be used to measure similarity between two states, we choose three of the representative embeddings to illustrate the performance of our model.

\begin{algorithm}[t]
\caption{\textsc{Rollout}} 
\label{alg:rollout}
\begin{algorithmic}[1]
\footnotesize
  \STATE \textbf{Input:} meta policy $\pi_{meta}$, low-level policy $\pi_{low}$.
  \STATE Initialize an episode $t \leftarrow 0$ and receive initial observation $o_0$ and a demonstration $\tau=\{o^\tau_1, o^\tau_2, \dots, o^\tau_T\}$.
  \WHILE{episode is not terminated} 
    \STATE Choose a sub-goal $g \sim \pi_{meta}(o_t, \tau)$ following \eqref{meta} and $t_0 \leftarrow t$
  	\WHILE{$\textit{is\_success}(o_{t}, o^\tau_g)$ is not true and episode is not terminated}
      \STATE $a_t \sim \pi_{low}(o_t, o^\tau_g)$
      \STATE $o_{t+1} \leftarrow \text{ENV}(o_t, a_t)$
      \STATE Store the transition $(o_t, o^\tau_g, a_t, o_{t+1})$ in low-level policy replay buffer $\mathcal{R}_\text{low}$.
      \STATE $t \leftarrow t + 1$
    \ENDWHILE
    \STATE Store the transition $(o_{t_0:t-1}, \tau, o^\tau_g, \textit{is\_success}(o_t, o^\tau_g), o_{t})$ in meta policy replay buffer $\mathcal{R}_\text{meta}$.
    \STATE Terminate if $g = T$ (\ie the agent reaches the end of the demonstration).
  \ENDWHILE
\end{algorithmic}
\end{algorithm}

\subsection{Selective Imitation Learning from Observations}
\label{sec:model}

To enable an agent to follow a new demonstration without action supervision, we propose a hierarchical RL framework that consists of two components: a meta policy and a low-level policy.
The meta policy chooses the sub-goal $o^\tau_g$ in the demonstration $\tau$.
Then, the low-level policy is executed to bring the current state closer to the sub-goal $o^\tau_g$ and is terminated when it is close enough to $o^\tau_g$.
This procedure is repeated to follow the demonstration (\figref{model} and \algref{rollout}).

We denote the meta policy as $\pi_{meta}(g|o_t,\tau;\theta)$ which is a neural network parameterized by $\theta$, where $o_t$ is the current observation, $\tau=\{o^\tau_1, o^\tau_2, \dots, o^\tau_T\}$ is a demonstration to imitate, and $g \in [1,T]$ is a demonstration state index of the sub-goal. 
The observation contains the robot joint configuration and position and orientation of the object. 
For efficient training, we constrain the input to the meta policy to $w$ future states $\{o^\tau_{i+1}, o^\tau_{i+2}, \dots, o^\tau_{i+w}\}$ in the demonstration $\tau$. 
As a result, the meta policy $\pi_{meta}(g|o_t,\tau;\theta)$ can be rewritten as following:
\begin{equation}
    g \sim \pi_\text{meta}(g|o_t, \{o^\tau_{i+1}, o^\tau_{i+2}, \dots, o^\tau_{i+w}\};\theta)= \text{argmax}_{g\in \{i+1, i+w\}} \gamma^{g-i-1}Q(o_t, o^\tau_g;\theta),
\label{eq:meta}
\end{equation}
where $o^\tau_i$ is the previous sub-goal state, $Q$ represents an action-value function of the meta policy, and $\gamma$ is the discounting factor to encourage the meta policy to not skip demonstration states.

Once the meta policy chooses a demonstration state $o^\tau_g$ as a sub-goal, the goal-conditioned low-level policy~\citep{schaul2015uvfa,mao2018universal} generates an action $a_t \sim \pi_{low}(a|o_t, o^\tau_g;\phi)$. 
The low-level policy conditioned on the sub-goal generates a rollout until the agent reaches the sub-goal (\ie $|o^\tau_g - o_{t+1}| < \epsilon$) or the episode ends. 
Note that action supervision for training the low-level policy is not available and the only reward signal comes from whether the agent reaches the sub-goal or not.
If the agent reaches the sub-goal, the meta policy picks the next sub-goal with respect to $o_{t+1}$. 
Also, the meta policy gets +1 reward for the transitions $(o_{t_0:t}, \tau, o^\tau_g, o_{t+1})$, which encourages the meta policy to maximize the number of achieved demonstration states while avoiding unreachable states. 
Otherwise, the meta policy gets 0 reward and the episode will be terminated. 
In this case, the meta policy learns to ignore unreachable demonstration states since it cannot get any further reward.
By repeating this process, the agent learns to follow only reachable states in the demonstration and take as many sub-goals as possible to maximize the reward for reaching sub-goals.

The proposed architecture has three advantages. First, the learner can take multiple low-level actions via the goal conditioned policy to accomplish a single action of the demonstrator. Hence, the learner can follow the expert demonstration even if the agent moves at a different temporal speed than the demonstrator due to proficiency or physical constraints. 
The next advantage is that the learner is able to deviate from the demonstration when some demonstration states are not reachable by the agent by skipping those states.
The third advantage is that using a meta policy to select a frame as a sub-goal from a long horizon helps an agent learn optimal behavior. 
If an agent follows demonstration states frame-by-frame, it can learn sub-optimal behavior due to the lack of action supervision. However, by learning to set a long-term goal that is reachable, the agent learns to reach further states instead of pursuing only short-term goals.



\subsection{Embedding}
\label{sec:embedding}

To check whether the low-level policy reaches the sub-goal, we compare the current observation $o_t$ and the selected demonstration state $o^\tau_g$. 
Since the current observation and the demonstration are collected by different agents in different environments, directly comparing pixel observations or robot configurations can be inaccurate. 
Therefore, we need an embedding space of these observations such that the distance between two embeddings implies the similarity between two observations with respect to the task (\ie the demonstration).

To apply an embedding to our framework, it needs to satisfy the following properties: (1) embeddings should contain all important information to judge progress of a task (\eg position of an object, interaction between an agent and an object) and (2) distance between two embeddings can represent how close the underlying states of the two observations are.
We use raw state, predicted 3D location, and AprilTag\footnote{AprilTag is a tagging library that gives position and orientation of detected tags (\url{https://april.eecs.umich.edu/software/apriltag}).} as object state representations. While our embeddings are simple, the framework can be applied to any embedding that satisfies these criteria.

\begin{figure}[t]
    \centering
    \includegraphics[width=.95\textwidth]{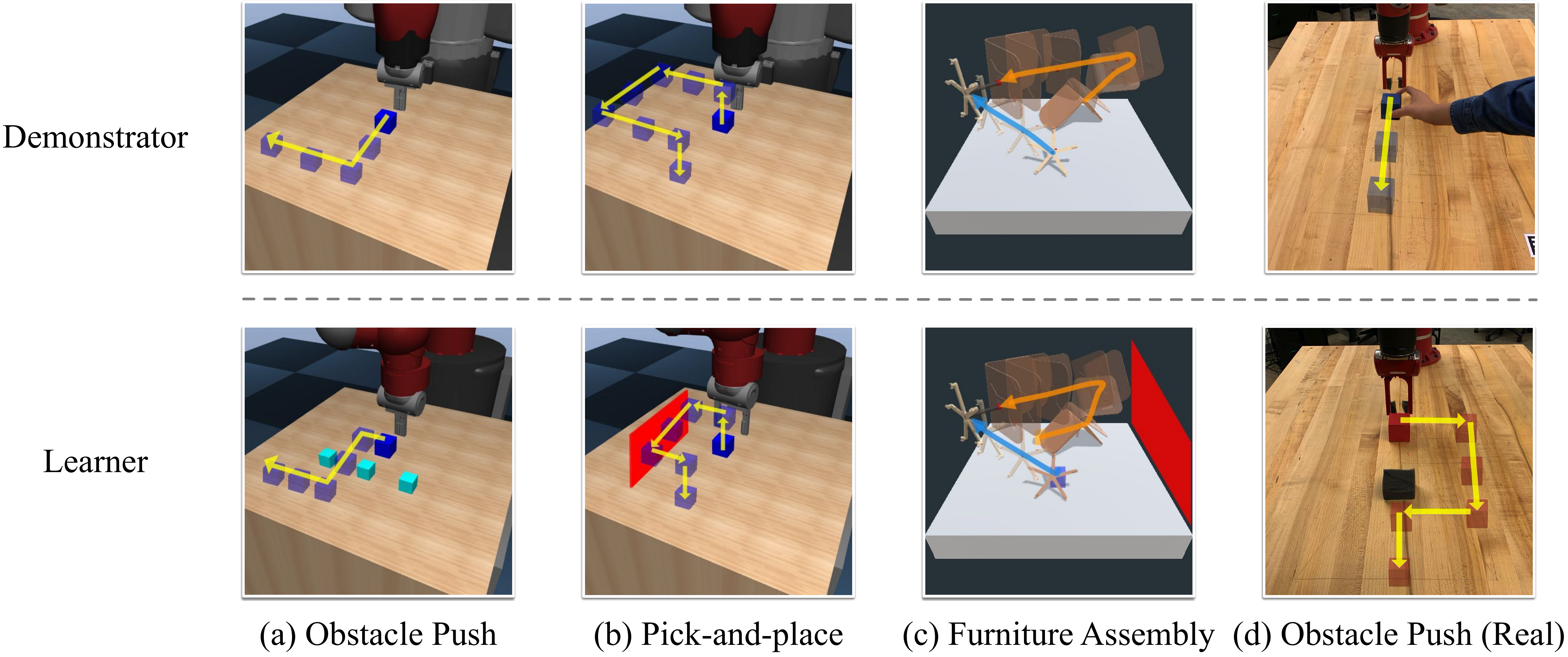}
    \caption{
    The top row shows demonstrations of the source agents (Baxter, Furniture movement, Human) and the bottom row shows execution with learners (Sawyer and Cursor).
    In (a) and (d), the demonstrators can freely move the block on the table while the learner has to deal with environment changes (an obstacle on the table). In (b) and (c), the learner must account for the physical limitation where the agent reach is bounded by red planes. Videos can be found on the website \url{https://clvrai.com/silo}.}
    \label{fig:task}
\end{figure}

\subsection{Training}
\label{sec:training}

We jointly train the meta policy and the low-level policy. 
Once an episode is collected following \algref{rollout}, we store meta policy transitions and low-level policy transitions into buffers $\mathcal{R}_\text{meta}$ and $\mathcal{R}_\text{low}$, respectively.
Then, we update the meta policy and low-level policy using samples from the corresponding buffers.
The meta policy is trained using double deep Q-network~\citep{DDQN} with a meta window size of 5 (\ie it picks the next sub-goal from the following 5 states in the demonstration). 
To train the low-level policy, we use soft actor-critic~\citep{haarnoja2018sac, haarnoja2018sac2}, which is an efficient off-policy model-free reinforcement learning algorithm.
Since our low-level policy is a goal-conditioned policy, we apply hindsight experience replay~\citep{andrychowicz2017hindsight} (HER) to improve data efficiency by augmenting virtual goals into collected transitions. Detailed description of training can be found in the supplementary material. At the beginning of the training phase, the meta policy cannot get rewards since the low-level policy cannot achieve any sub-goals.
However, the low-level policy can still learn to achieve many different goals visited during random exploration using HER~\citep{andrychowicz2017hindsight}.

\section{Experiments}

\begin{figure}[t]
    \centering
    \begin{subfigure}[t]{0.3\textwidth}
    	\centering
    	\includegraphics[width=\textwidth]{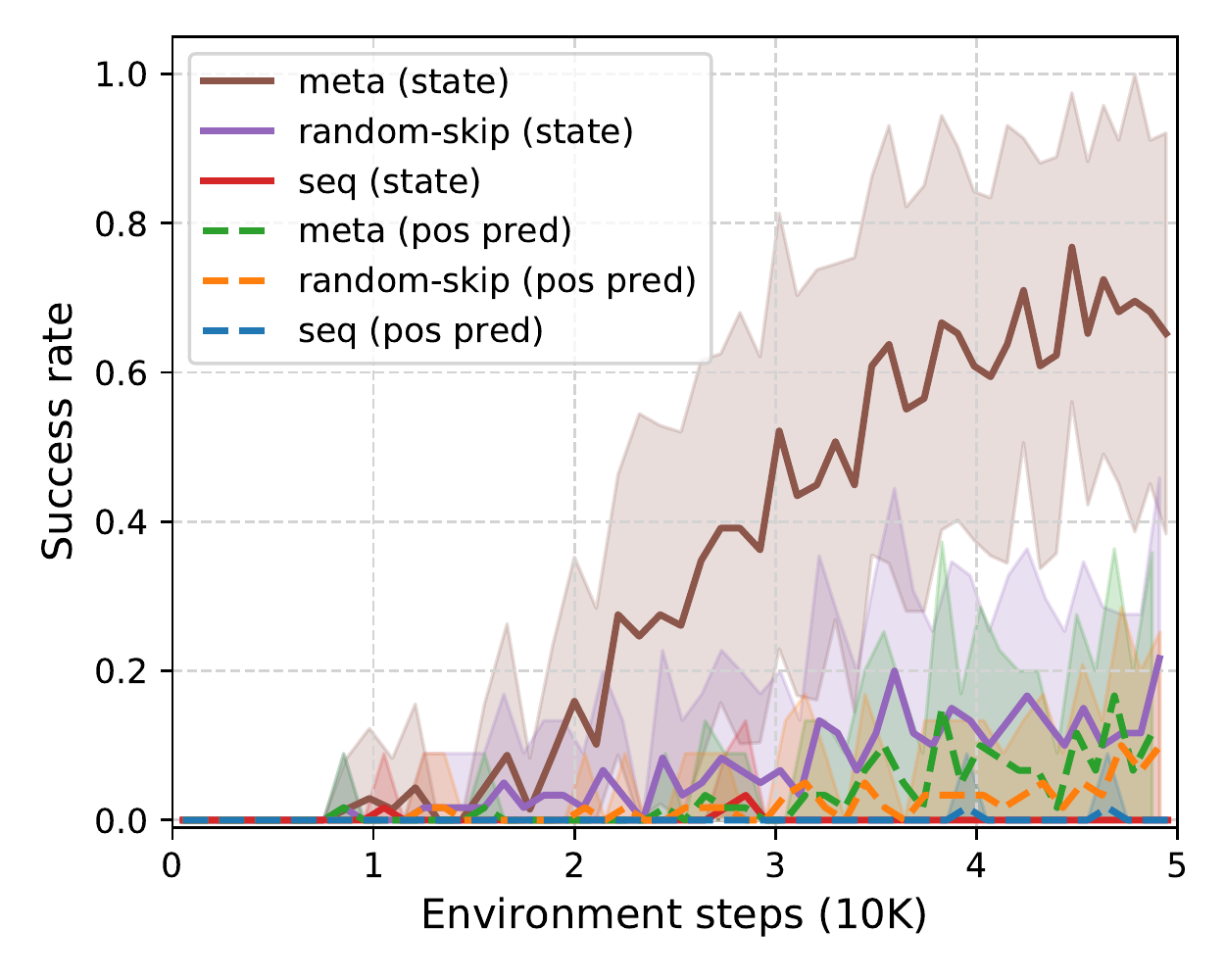}
        \caption{\textsc{Obstacle push}}\label{fig:result_push}
    \end{subfigure}
    \quad
    \begin{subfigure}[t]{0.3\textwidth}
    	\centering
    	\includegraphics[width=\textwidth]{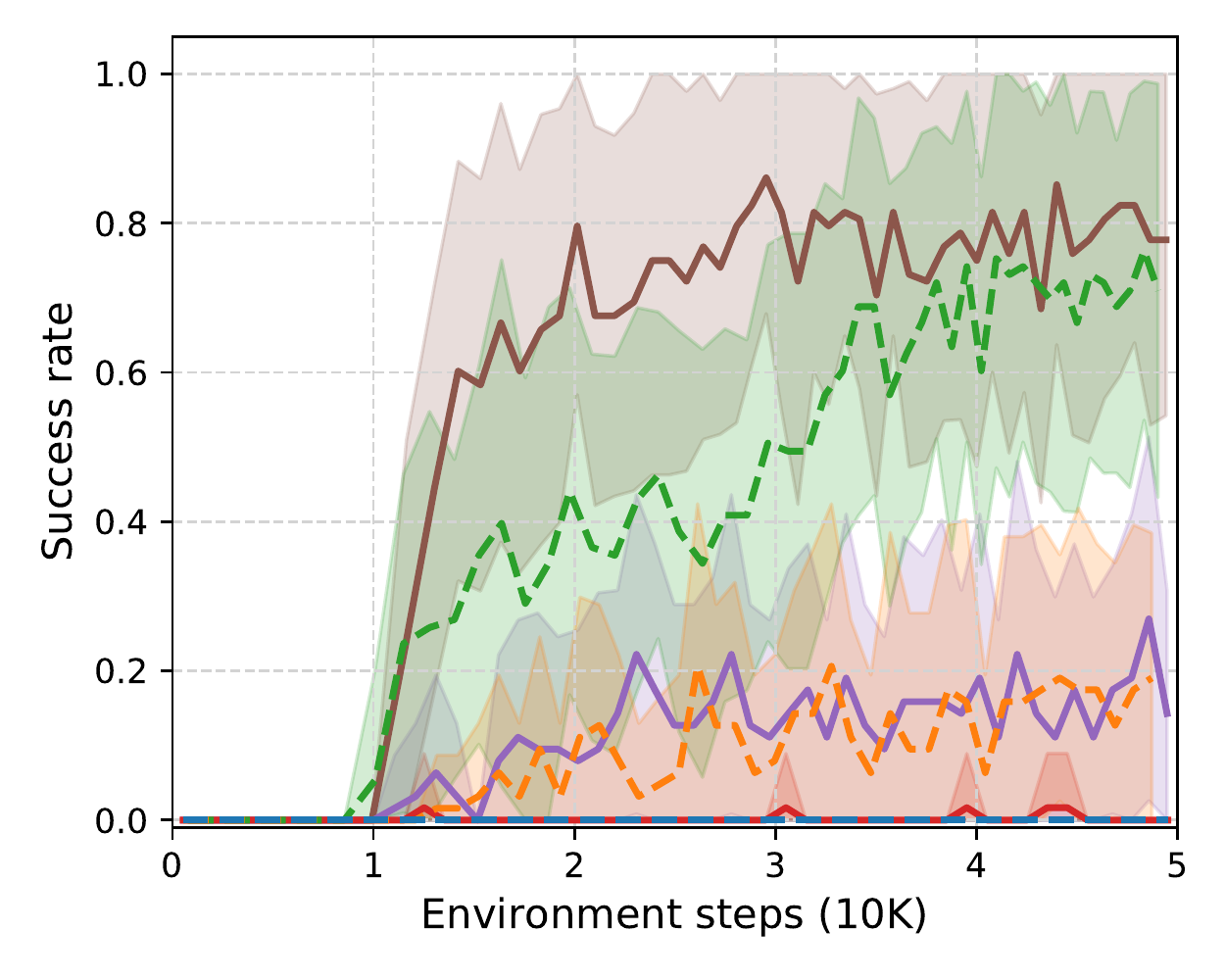}
        \caption{\textsc{Pick-and-place}}\label{fig:result_pick_place}
    \end{subfigure}
    \quad
    \begin{subfigure}[t]{0.3\textwidth}
    	\centering
    	\includegraphics[width=\textwidth]{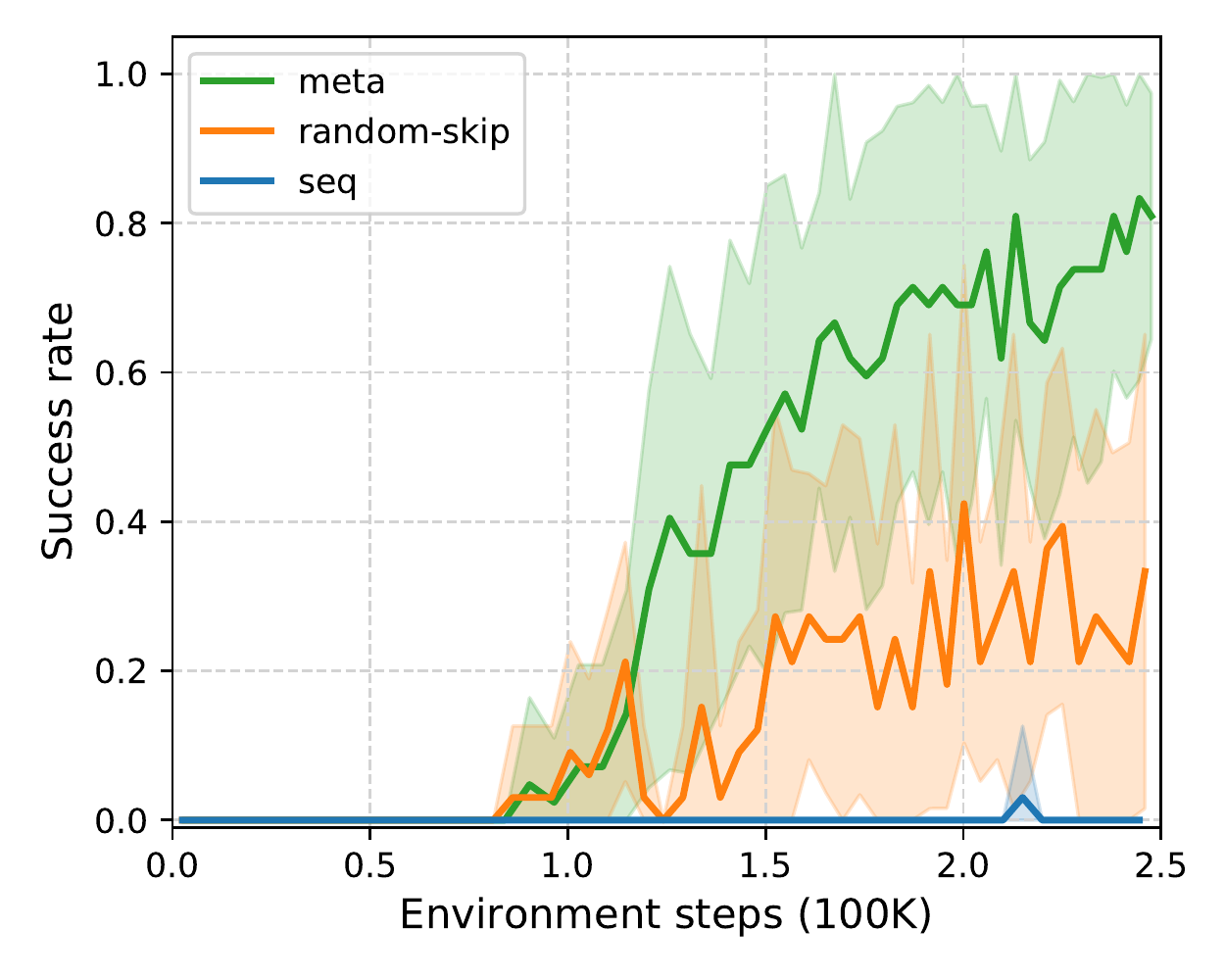}
        \caption{\textsc{Furniture assembly}}\label{fig:result_furniture}
    \end{subfigure}
    \caption{
        Success rates on \textsc{Obstacle push}, \textsc{Pick-and-place}, and \textsc{Furniture assembly} environments. Our approach outperforms baselines in all tasks. The \textit{sequential} baseline could not solve any task due to the unreachable states in demonstrations. The \textit{random-skip} baseline achieves around $0.2$ success rate since it can avoid unreachable states by chance. We also evaluate our method with the object location predictor (pos pred) which shows comparable results to the raw state inputs in \textsc{Pick-and-place} but performs poorly on \textsc{Obstacle push}.
    }
    \label{fig:result_simulated}
\end{figure}

We evaluate our method to show how a meta policy that explicitly picks feasible sub-goals benefits imitation learning from unaligned demonstrations.
Especially, we are interested in answering two questions: (1) how closely can SILO match the demonstration? and (2) can it reach the end frame of the demonstration? 
To answer these questions and to illustrate the potential of SILO, we design tasks that require both imitation and exploration by putting an obstacle in the learner's trajectory or limiting the learner's physical capability as shown in \figref{task}, so that a learner has to selectively imitate demonstrations. 
We conduct experiments in both simulated and real-world tasks: 
\begin{itemize}[leftmargin=0.3in]
    \item \textsc{Obstacle push}: push a block so that it follows a trajectory described in a demonstration. An obstacle is placed during training time to make step-by-step imitation impossible.
    \item \textsc{Pick-and-place}: pick up a block and move it following a trajectory described by a demonstration. The learner has shorter reach than the demonstrator. Hence, it needs to ignore the out-of-reach motions and imitate only the reachable frames.
    \item \textsc{Furniture assembly}: assemble furniture by aligning two parts. The demonstration describes how to move and align furniture pieces to assemble but is missing an agent. The environment contains two cursors and the cursors can assemble furniture by repeating: picking two pieces, moving them toward each other, and aligning two pieces. The agent requires 3D understanding including 3D positions/rotations and their physical interactions. 
    \item \textsc{Obstacle push (Real)}: push a block to a target position following a human demonstration in the real world. During training time, we put an obstacle in the middle of the table so that the Sawyer robot has to skip some parts of the demonstration.
\end{itemize}

We use a simulated Sawyer robot for \textsc{Obstacle push} and \textsc{Pick-and-place} and a real Sawyer robot for \textsc{Obstacle push (real)}. 
To cover several challenges including 3D alignment of two objects, we used the IKEA furniture assembly environment~\citep{lee2019ikea} that supports assembling IKEA furniture models.
The simulated environments are developed based on the Stanford Surreal environment~\cite{fan2018surreal} and simulated in the Mujoco~\cite{todorov2012mujoco} physics engine.
For each simulated environment and method, a model is trained with three different random seeds.
Details about observations, actions, and demonstrations can be found in the supplementary. Videos can be found on the website \footnote{\url{https://clvrai.com/silo}}.

We tested three state representations: raw state and object location prediction for simulated tasks, and AprilTag for real-world experiments. 
For an object location prediction, we use a CNN to predict the 3D coordinate of the object from raw pixels. The agent is considered to reach a sub-goal when the current state and the sub-goal are within 2 cm in the Euclidean distance.
In real-world experiments, we use AprilTag for the embedding and set 2.5 cm as a threshold to determine sub-goal success.

\subsection{Simulated Tasks}

To investigate the need of the meta policy to selectively follow demonstrations, we compare our learned meta policy (\textit{meta}) with sequential execution (\textit{seq}) and random skip (\textit{random-skip}) baselines. 
The sequential baseline always selects the next frame in the demonstration as the sub-goal. It serves as a benchmark of the difficulty in following demonstrations frame-by-frame.
The random-skip baseline picks a random frame in the meta window as the next sub-goal and it shows how beneficial a learned meta policy is over avoiding unreachable states by random chance. We use 20 demonstrations to train our model for simulated tasks.

\begin{figure}[t]
    \centering
    \includegraphics[width=1\textwidth]{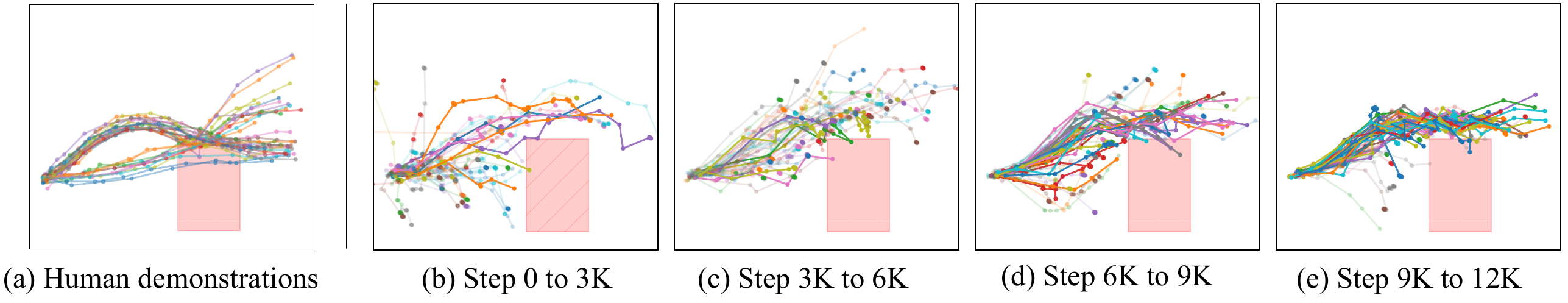}
    \caption{
            (a) shows $(x,y)$-trajectories of the human demonstrations for \textsc{Obstacle Push (Real)}. All trajectories go through the obstacle (red box), making it infeasible to strictly follow any of them. (b)-(e) display the trajectories of the block over training steps. Trajectories where the low-level policy reaches the meta policy's goal are bolded. Over time, the meta policy learns to pick sub-goals that are away from the obstacle and the low-level policy learns to realize the sub-goals.
    }
\label{fig:trajectorytime}
\end{figure}

\textbf{Obstacle Push:} In \textsc{Obstacle push}, the learner must learn to deviate from the demonstration trajectory to avoid the obstacle, and go back to the demonstration trajectory as soon as possible. 
\figref{result_simulated} shows that sequentially following demonstration could not succeed because all demonstrations are blocked by the obstacle. The random-skip baseline managed to succeed in a few trials as it skips the unreachable states in demonstration by chance. However, its success rate is bounded by a small threshold, $0.23$, since there is a high chance to pick an unreachable state as a sub-goal. Our meta policy outperforms both baselines with a significant margin as it learns the limitation of the agent and environment to select suitable sub-goals. 
The experiments with the object location prediction could not follow demonstration mainly because the prediction error near the obstacle makes an agent not be able to reach sub-goals around the obstacle.

\textbf{Pick-and-place:}
The optimal strategy for \textsc{Pick-and-place} is to skip the states that are unreachable and directly move to the first future state within its action space.
Both sequential and random-skip baselines struggle with low success rate as demonstrated in \figref{result_simulated}. 96\% of the failures (289/300 for sequential and 238/245 for random-skip) are due to choosing demonstration states not reachable by an agent (\ie the block in the selected demonstration states are outside of Sawyer's range) and 4\% of them are due to the failure of the low-level policy reaching the chosen sub-goals. However, our meta policy achieves a clear success (95\% success rate) in this task. \figref{result_simulated}b demonstrates that our method with the predicted object location achieves similar results with ones with the raw state, which shows the robustness of our framework to noise in state embeddings.

The \textsc{Pick-and-place} experiments show that the meta policy enables the agent learn quickly because it prevents a policy from a local minimum. We observe that the agent without the meta policy learns to push the block into the table, instead of picking it up. This leads the block to pop out so that it reaches the second demonstration frame where the block is slightly elevated. This behavior is easy to learn, but not correct behavior for picking the block.

\textbf{Furniture Assembly:}
The demonstration shows how to move and align two parts to attach them together, without any cursors. Then, two cursors learn to follow the demonstration. Since the furniture model can move outside the cursor's action space in demonstrations, an agent requires to deviate from demonstrations. Our method effectively learns to ignore those demonstration states and follow feasible parts of the demonstrations while baselines are unable to.

\begin{table}[h]
    \centering
    \caption{
    	Success rate for all tasks, comparing our method against sequential and random-skip baselines. 
        Each entry in the table represents average success rate over 100 held-out demonstrations for simulated environments and 20 demonstrations for the real-world environment. Success rate is defined by reaching the end of the demonstration frame. The number in parenthesis represents the coverage which is percentage of demonstration frames fulfilled by the agent.
        }
    \vspace{5pt}
    \scalebox{0.9}{
    \begin{tabular}{ccccc}
        & Obstacle push & Pick-and-place & Furniture assembly & Obstacle push (Real) \\
        \hline
        Sequential & 
        0.00 (0.424) & 0.00 (0.676) & 
        0.00 (0.30) & N/A	 \\    

        Random-skip & 
        0.13 (0.173) & 0.23 (0.567) 
        & 0.10 (0.20) & 0.1 (0.195)	\\ 

        Meta policy (ours) &         
        \textbf{0.70 (0.496)} & \textbf{0.95 (0.811)}
        & \textbf{0.40 (0.41)} & \textbf{0.25 (0.496)}	\\ 
    \end{tabular}
    }
    \label{tab:evaluation}
\end{table}

\subsection{Obstacle Push in the Real World}

The task requires the Sawyer to push a block towards a desired location in the real world. 
We first collect 32 training demonstrations by moving a box toward a randomly chosen goal using a human hand (\figref{trajectorytime}a). After the demonstrations are collected, another block is glued on the table to act as an obstacle and it overlaps with all demonstration trajectories.
As shown in \figref{trajectorytime}e, our method learns to push around the obstacle by skipping the unachievable portions of the demonstration where the block is in the obstacle area. 
On the other hand, the random-skip baseline is unable to consistently achieve this because it cannot bypass the states in the demonstration where the block is moving through the obstacle area. 
We do not consider the sequential baseline as all demonstrations are blocked by the obstacle therefore making sequential execution impossible.
As seen in \tabref{evaluation}, SILO achieves 0.25 success rate on 20 test demonstrations while the random-skip baseline achieves 0.1 success rate after 12-hours of training which corresponds to 11,000 environment steps. 
Due to the low sample size, the policy learns to succeed on demonstrations with ending frames further away from the obstacle because there is reduced likelihood of getting stuck to the obstacle. However, with more training time, we expect the policy to generalize to the full goal distribution as seen in the simulated experiments.

\begin{figure*}[t]
\centering
    \begin{subfigure}[t]{0.3\textwidth}
    	\centering
    	\includegraphics[width=\textwidth]{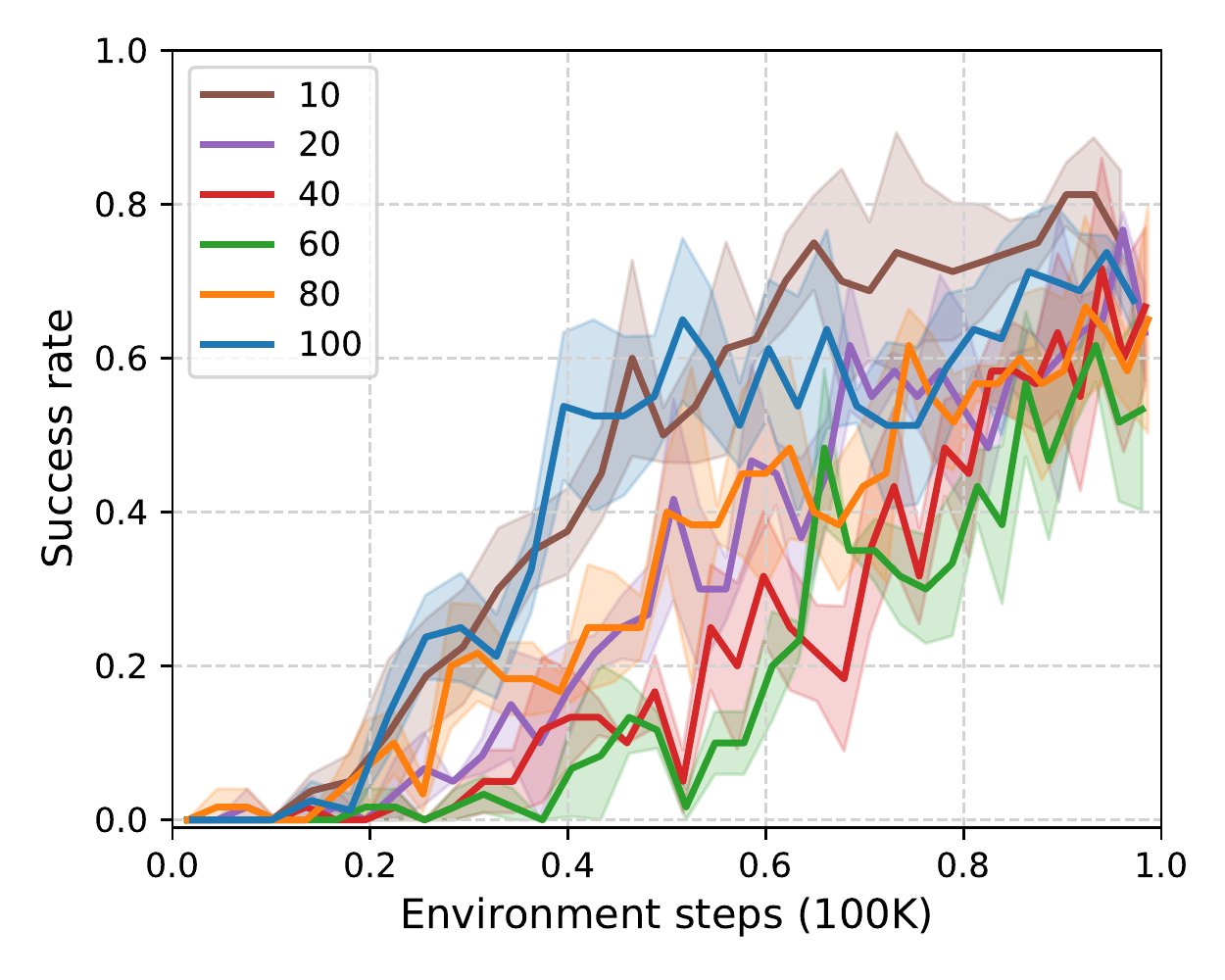}
        \caption{\textsc{Obstacle push}}
        \label{fig:result_push_num_demo}
    \end{subfigure}
    \quad
    \begin{subfigure}[t]{0.3\textwidth}
    	\centering
    	\includegraphics[width=\textwidth]{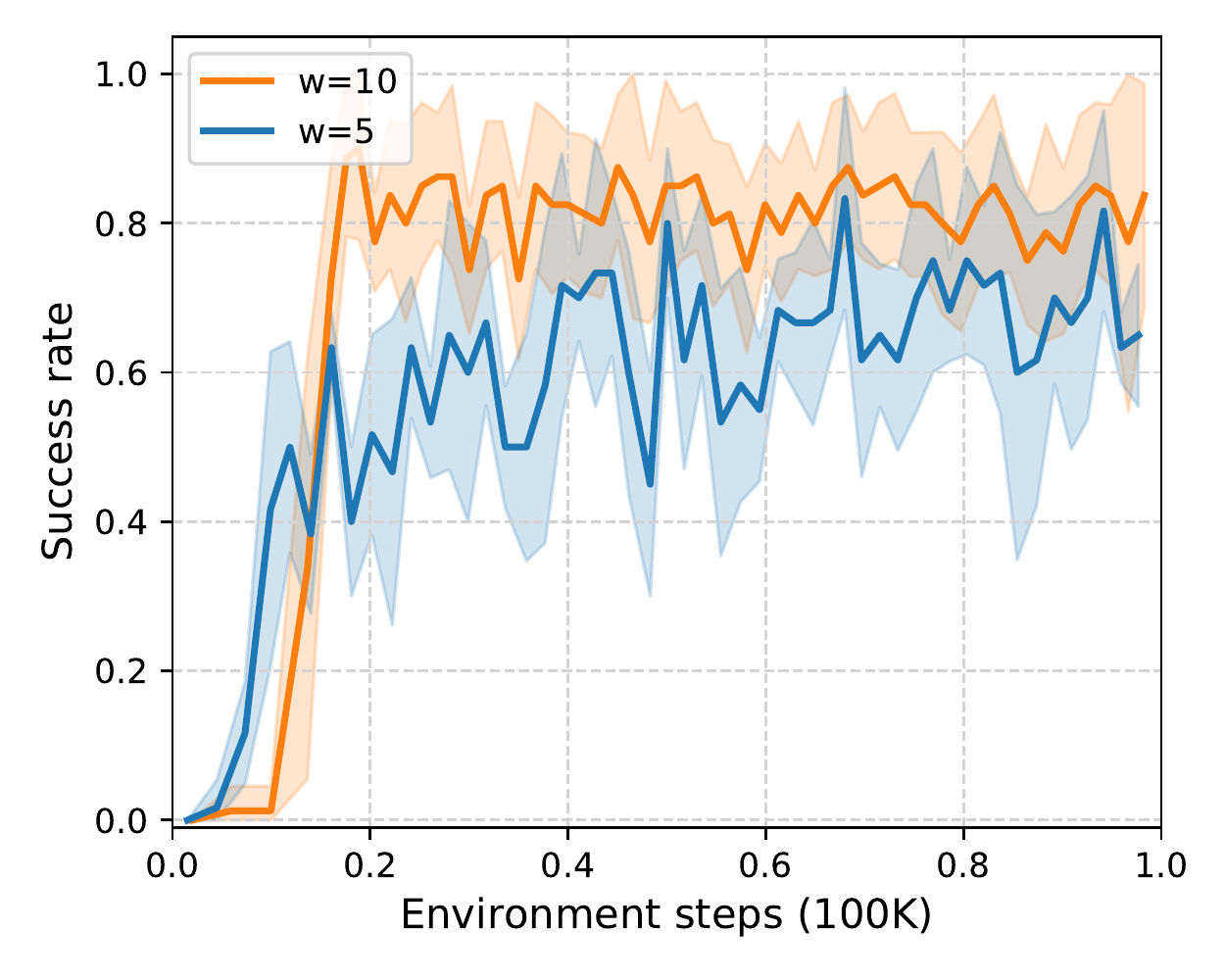}
        \caption{\textsc{Pick-and-place}}
        \label{fig:result_pick_place_meta_window}
    \end{subfigure}
    \caption{
        (Left) Our model is trained with 10, 20, 40, 60, 80, and 100 demonstrations in \textsc{Obstacle push}. 
        (Right) Our model is trained with meta window 5 and 10 in \textsc{Pick-and-place}. 
    }
\end{figure*}

\subsection{Ablation Study}

\textbf{Number of training demonstrations: } The experiments show that SILO learns to selectively imitate the given demonstration. We further evaluate which factor affects the performance of our method. First, we train our method on \textsc{Obstacle push} with 10, 20, 40, 60, 80, and 100 demonstrations and evaluate on another 100 held-out demonstrations. 
Our method can learn to imitate demonstrations only with 10 demonstrations (0.87 success rate) and models with different number of training demonstrations show marginal differences in the learning curves and the evaluation results (see \figref{result_push_num_demo}). SILO learns to follow an unseen demonstration since the low-level policy experiences diverse states with the entropy maximization objective~\citep{haarnoja2018sac} and is trained to reach any states it has once visited using HER~\citep{andrychowicz2017hindsight}.

\textbf{Size of meta window: } We evaluate our method on \textsc{Pick-and-place} with reduced reach of Sawyer, which requires to skip 5-9 demonstration states, whereas the original experiment requires to skip 1-4  states with the meta window size of 5. To deal with this large number of frames to skip, we set the meta window size to 10. The results in \figref{result_pick_place_meta_window} show similar training curves, but shifted to the right 2K environment steps as the meta policy needs more exploration. This shows that our method can efficiently learn to bypass the large number of infeasible demonstration states.
\section{Conclusion}

We present selective imitation learning from observations (SILO) as a step towards learning from diverse demonstrations. Our method selects reachable states in the demonstration and learns how to reach the selected states. Since the underlying actions to perform a demonstration are not required, SILO enables a robot to learn from humans and other agents. Additionally, the agent does not need to follow the demonstration step-by-step, increasing transfer and generalization capabilities. To evaluate SILO, we propose 3 environments, \textsc{Obstacle Push}, \textsc{Pick-and-place}, and \textsc{Furniture Assembly} for imitation learning.
The experiments demonstrate that SILO outperforms baselines and can be deployed in real robots. We also show that SILO can work on estimated input and goal states. SILO can serve as a natural starting point for utilizing vision-based embeddings to learn more complex behaviors and interactions. 
Viewpoint agnostic embeddings~\citep{sermanet2018time}, perceptual reward functions~\citep{sermanet2017rewards}, goal distance estimation~\citep{yuDPN, srinivas2018universal}, and reward learning from human feedback~\citep{singh2019end-to-end} could be applied with our framework as an exciting direction for future work.



\clearpage
\acknowledgments{
This project was funded by the center for super intelligence, Kakao Brain, and SKT. 
We would like to thank Utkarsh Patel, Hejia Zhang, Avnish Narayan, and Naghmeh Zamani for advice and support on a Sawyer robot. We also thank Shao-Hua Sun, Karl Pertsch, Ayush Jain, and Jingyun Yang for important suggestions and discussion.
}


\bibliography{bib/drl,bib/detection,bib/deep_learning,bib/rl,bib/hrl,bib/env,bib/gnn,bib/sim2real,bib/robotics,bib/goal_directed_rl,bib/meta,bib/imitation}

\clearpage
\appendix

\section{Environment details}

The details of observation spaces, action spaces, and episode lengths are described in Table~\ref{table:env_detail}. Note that all measures are in meters, and we omit the measures here for clarity of the presentation.

No explicit reward is provided for \textsc{Obstacle push}, \textsc{Pick-and-place}, and \textsc{Furniture assembly}. The intrinsic reward for the meta policy is whether the low-level policy reaches the sub-goal chosen by the meta policy. Hence, the return of an episode is defined by number of sub-goals reached throughout the episode.

\subsection{Obstacle Push}

In the simulated pushing experiment, a demonstration shows a Baxter robot pushing a block to the target position. 
Then, the learner, a Sawyer robot arm, is asked to push the block to the same target position following this demonstration. 
However, the environment that Sawyer is facing is different from the one Baxter has faced in the demonstrations: we place an obstacle that blocks all trajectories of the block in the demonstrations. Thus, in order to maximize the reward, the learner must learn to deviate from the demonstration trajectory to avoid the obstacle, and go back to the demonstration trajectory as soon as possible.

\textbf{Observations}: The observation space consists of the block $(x, y, z)$ coordinates and rotation $(w, x, y, z)$ as well as the robot configuration including the gripper position, rotation, velocity, and angular velocity.

\textbf{Actions}: The action space is relative offset $dx, dy$ of the $(x, y)$ position of the gripper. In \textsc{Obstacle push} task, the gripper cannot move vertically.

\textbf{Initialization}: 
In simulated manipulation tasks, a block is randomly placed in a circle centered at the table center and with a radius of 0.02 .
A robot arm is initialized to place its gripper behind the block.
The obstacle of size $0.04 \times 0.04 \times 0.02$ is placed during training 0.16 m ahead of the block. The trajectories described in the demonstrations always pass through the obstacle. The goal position is $(x, y)$ where $(x, y) = (0.25, 0)$ with randomness of $0.1$.

\subsection{Pick-and-place}
Pick-and-place~\cite{duan2017one-shot,xu2018NTP} is a well-known manipulation task in the imitation learning field since it requires manipulation skills and understanding of composition of the task structure. 
In the pick-and-place experiment, we limit the horizontal action space of the learner, a Sawyer robot arm, to be smaller than the demonstrator, a Baxter robot. 
The Baxter picks up a block and move it to the target area with a C-shape trajectory. During the first stage, Baxter moves the block in left or right direction until the block is out of the action space of Sawyer. Then, Baxter moves the block to the target area. 

\textbf{Observations}: The observation space consists of the block $(x, y, z)$ coordinates and rotation $(w, x, y, z)$ as well as the robot configuration including the gripper position, rotation, velocity, and angular velocity.

\textbf{Actions}: The action space is relative offset $dx, dy, dz$ of the $(x, y, z)$ position of the gripper.

\textbf{Initialization}: 
A block is placed in the center of the table and a robot arm is initialized to place its gripper above the block.
The demonstrations show a C-shape curve which passes two milestones. The first one is on the left/right side of the table where $\text{abs}(x) > 0.14$ and the height of it is sampled in $[0.06, 0.07]$. Another one is located in the center area where $x \sim \text{uniform}(-0.05, 0.05)$ and $z \sim \text{uniform}(0.06, 0.07)$.

\subsection{Furniture assembly}

To make the problem of furniture assembly feasible, we explicitly split the task into two phases: move one part to the center of the scene and then align another part with the first one. 
Hence, we need to manipulate only one cursor and consider one object at a time.

\textbf{Observations}: Observation for each time-step consists of the position $(x, y, z)$ and orientation $(w, x, y, z)$ of a target object in the scene, the $(x, y, z)$ coordinate of the target cursor, and whether the cursor is grasping a piece or not. 

\textbf{Actions}: The action space is relative offset $dx, dy, dz$ of the $(x,y, z)$ position of the cursor, relative rotation $(rx, ry, rz)$, and gripper state $dg$.

\subsection{Obstacle Push (Real)}

\textbf{Observations}: In \textsc{Obstacle Push (Real)}, the observation space consists of the block $(x,y)$ coordinates and rotation $\sin{\theta}, \cos{\theta}$ relative to the robot base as well as the $(x,y)$ coordinate and rotation of the gripper.

\textbf{Actions}: The action space is relative offset $dx, dy, d\theta$ of the $(x,y)$ position of the gripper as well as the rotation of the gripper. The rotation is limited to 30 degrees.

\textbf{Initialization}: The box is initialized to a central position. The gripper is initialized 3 cm behind the box with 2 cm of noise added in the tangent direction. The obstacle is placed 15 centimeters in front of the box, and is 4 cm by 5 cm. The goal positions lie 30 cm in front of the box starting position and vary in the tangent direction. 

\textbf{Data Collection}:
To collect the demonstrations, we provide a script to capture the block positions over time. Specifically, the script uses AprilTag to record the given tag coordinates. The block is manually moved by a human while the script is running. Each demonstration contains 10-15 time-steps of the block position as the block is pushed from start to goal. We recorded pushes varying in intermediate trajectory movement as well as ending position. 

\textbf{Rewards}: Due to the usage of negative reward with HER, care must be taken to penalize early termination conditions. Otherwise, the agent may seek a premature termination to minimize its overall negative reward. In the case of camera occlusion or out of boundary block positions, the episode is terminated and a negative penalty of $t_{max} - t_{alive}$ is applied where $t_{max}$ is the maximum number of steps in an episode.

\begin{table}[ht]
  \caption{Environment details}
  \label{table:env_detail}
  \centering
  \begin{tabular}{lcccc}
    & {Obstacle push} & {Pick-and-place} & {Furniture} & {Obstacle push (real)} \\
    \hline
    Observation Space & 21 & 21 & 10 & 7 \\
    - Robot observation & 14 & 14 & 3 & 3 \\
    - Block observation & 7 & 7 & 7 & 4 \\
    \hline
    Action Space & 2 & 4 & 7 & 3 \\
    - Move Actions & 2 & 3 & 3 & 2 \\
    - Rotate Actions & 0 & 0 & 3 & 1 \\
    - Gripper Action & 0 & 1 & 1 & 0 \\
    \hline
    \# Demonstrations & 20 & 20 & 20 & 32 \\
    Episode length & 50 & 50 & 150 & 30 \\
  \end{tabular}
\end{table}

\section{Training details}

We use PyTorch~\citep{paszke2017pytorch} for our implementation and all experiments are conducted on a workstation with Intel Xeon E5-2640 v4 CPU and a NVIDIA Titan Xp GPU.

\subsection{Networks}

The meta policy is represented as a Q-network $Q^{\pi_\text{meta}}(o_t, o^\tau_g;\theta)$. The low-level policy  $\pi_\text{low}(a|o_t, o^\tau_g;\phi)$ is trained with two critic networks $Q^{\pi_\text{low}}_1(o_t, o^\tau_g,a_t;\phi)$ and $Q^{\pi_\text{low}}_2(o_t, o^\tau_g,a_t;\phi)$ as suggested in \citet{haarnoja2018sac}. 
Both the meta policy and the low-level policy take as input the current observation $o_t$ and $o^\tau_g$.
The meta policy outputs a Q-value while the low-level policy outputs the mean and standard deviation of a Gaussian distribution over an action space. For low-level policy, we apply $\tanh$ activation to normalize the action.
All networks in this paper consist of 2 fully connected layers of 128 hidden units with ReLU nonlinearities.

\subsection{Hyperparameters}

For all networks, we use the Adam optimizer~\citep{kingma2014adam} with mini-batch size of 256 and learning rate 3e-4. After each rollout, we update the networks 50 times. We limit the maximum length of an episode as 50, 50, 150, and 30 for \textsc{Obstacle push}, \textsc{Pick-and-place}, \textsc{Furniture}, and \textsc{Obstacle push (real)}, respectively.
\tabref{hyperparameter} lists the parameters used across all environments.

\begin{table}[ht]
  \caption{Hyperparameters}
  \label{tab:hyperparameter}
  \centering
  \begin{tabular}{cc}
    Parameters & Value \\
    \hline
    learning rate & 3e-4 \\ 
    meta window & 5 \\
    meta reward decay & 0.99 \\
    gradient steps & 50 \\
    batch size & 256 \\
    discount factor & 0.99 \\
    target smoothing coefficient & 0.005 \\
    epsilon decay & 0.005 \\
    reward scale (SAC) & 1.0 \\
    experience buffer size & 1e5 \\

  \end{tabular}
\end{table}

\subsection{Experience Replay Buffer}
We use two experience replay buffers, one for the meta policy $\mathcal{R}_\text{meta}$ and another one for the low-level policy $\mathcal{R}_\text{low}$.
Each replay buffer can store up to 1e5 episodes.
When we sample batch from $\mathcal{R}_\text{low}$, we augment a virtual goal using HER~\citep{andrychowicz2017hindsight} with 0.8 probability.

\subsection{3D Object Position Prediction}

We tested our method with the predicted 3D position of an object, not with the ground truth state.
We pre-train a 3D position prediction network by collecting 50,000 pairs of randomly generated visual frame and object position in \textsc{Obstacle push} and \textsc{Pick-and-place}.
The network consists of 3 convolutional layers with kernel size 3, stride 2, and 64 channels, followed by two fully-connected layers with 128 hidden units and outputs 3 numbers representing $(x, y, z)$ coordinate of the object. 
The size of visual input is $128\times128\times3$ and we use ReLU as a nonlinearity activation.
We use the Adam optimizer~\citep{kingma2014adam} with mini-batch size of 4 and learning rate 1e-3 to minimize the mean squared error of the predicted coordinates.
We trained the network for 40,000 updates and the mean squared loss converges to 9.765e-6 which corresponds to 3 mm errors.

\begin{figure}[h]
    \centering
	\centering
	\includegraphics[width=\textwidth]{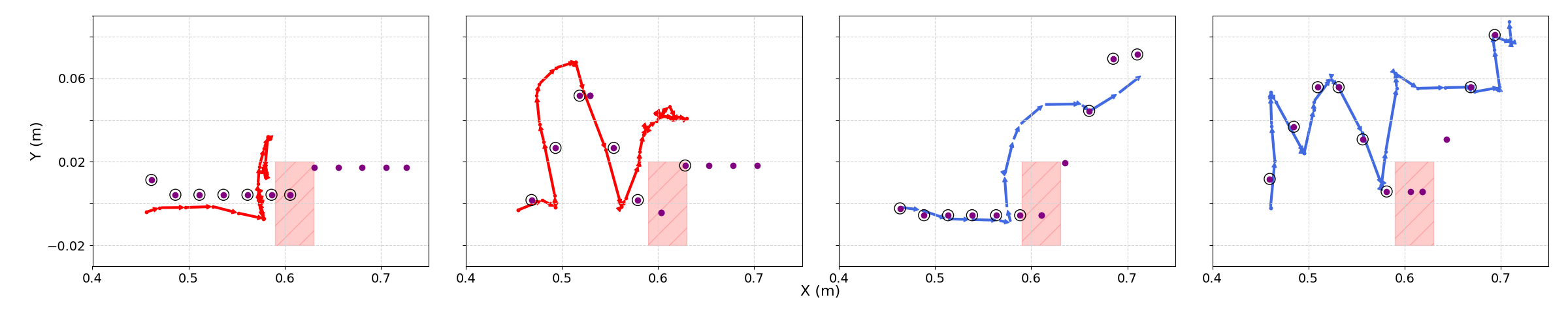}
  
    \caption{Visualization of \textsc{Obstacle push (real)} demonstrations and learner's trajectories. The purple dots represent demonstration frames and circled purple dots represent frames chosen by the policy. Blue trajectories (right) are successful while red trajectories (left) are not. Choosing goals within the obstacle leads to failure while learning to choose goals outside the obstacle leads to success.}
    \label{fig:trajectory_push}
\end{figure}

\end{document}